\documentclass{article}

\PassOptionsToPackage{numbers}{natbib}

\usepackage[preprint]{neurips_2019}

\usepackage[utf8]{inputenc} 
\usepackage[T1]{fontenc}    
\usepackage{hyperref}       
\usepackage{url}            
\usepackage{booktabs}       
\usepackage{amsfonts}       
\usepackage{nicefrac}       
\usepackage{microtype}      

\usepackage{amsmath}
\usepackage{graphicx}
\usepackage{multirow}
\usepackage{multicol}
\usepackage{threeparttable}

\title{Neural Logic Networks}

\author{%
Shaoyun Shi$^1$\thanks{This work was done when the first author was visiting at Rutgers University.},~Hanxiong Chen$^2$,~Min Zhang$^1$,~Yongfeng Zhang$^2$\thanks{Corresponding author.} \\
$^1$Tsinghua University, Beijing, China~$^2$Rutgers University, NJ, USA \\
\texttt{\{shisy17,z-m\}@tsinghua.edu.cn, \{hc691,yongfeng.zhang\}@rutgers.edu} \\
}

\begin{document}

\maketitle

\begin{abstract}
Recent years have witnessed the great success of deep neural networks in many research areas. 
The fundamental idea behind the design of most neural networks is to learn similarity patterns from data for prediction and inference, which lacks the ability of logical reasoning. However, the concrete ability of logical reasoning is critical to many theoretical and practical problems.
In this paper, we propose Neural Logic Network (NLN), which is a dynamic neural architecture that builds the computational graph according to input logical expressions. It learns basic logical operations as neural modules, and conducts propositional logical reasoning through the network for inference. Experiments on simulated data show that NLN achieves significant performance on solving logical equations. Further experiments on real-world data show that NLN significantly outperforms state-of-the-art models on collaborative filtering and personalized recommendation tasks.
\end{abstract}
    
\section{Introduction}

Deep neural networks have shown remarkable success in many fields such as computer vision, natural language processing, information retrieval, and data mining. 
The design philosophy of most neural network architectures is learning statistical similarity patterns from large scale training data. For example, representation learning approaches learn vector representations from image or text for prediction, while metric learning approaches learn similarity functions for matching and inference. Though they usually have good generalization ability on similarly distributed new data, the design philosophy of these approaches makes it difficult for neural networks to conduct logical reasoning in many theoretical or practical tasks.


However, logical reasoning is an important ability of human intelligence, and it is critical to many theoretical problems such as solving logical equations, as well as practical tasks such as medical decision support systems, legal assistants, and collaborative reasoning in personalized recommender systems. In fact, logical inference based on symbolic reasoning was the dominant approach to AI before the emerging of machine learning approaches, and it served as the underpinning of many expert systems in Good Old Fashioned AI (GOFAI). However, traditional symbolic reasoning methods for logical inference are mostly hard rule-based reasoning, which may require significant manual efforts in rule development, and may only have very limited generalization ability to unseen data.


To integrate the advantages of deep neural networks and logical reasoning, we propose Neural Logic Network (NLN), a neural architecture to conduct logical inference based on neural networks. NLN adopts vectors to represent logic variables, and each basic logic operation (AND/OR/NOT) is learned as a neural module based on logic regularization. 
Since logic expressions that consist of the same set of variables may have completely different logical structures, capturing the structure information of logical expressions is critical to logical reasoning. To solve the problem, NLN dynamically constructs its neural architecture according to the input logical expression, which is different from many other neural networks. By encoding logical structure information in neural architecture, NLN can flexibly process an exponential amount of logical expressions. 

Extensive experiments on both theoretical problems such as solving logical equations and practical problems such as personalized recommendation verified the superior performance of NLN compared with state-of-the-art methods.




\section{Neural Logic Networks}
\label{sec:nln}

Most neural networks are developed based on fixed neural architectures, either manually designed or learned through neural architecture search. Differently, the computational graph in our Neural Logic Network (NLN) is built dynamically according to the input logical expression. 
In NLN, variables in the logic expressions are represented as vectors, and each basic logic operation is learned as a neural module during the training process. We further leverage logic regularizers over the neural modules to guarantee that each module conducts the expected logical operation.

\subsection{Logic Operations as Neural Modules}
An expression of propositional logic consists of logic constants (T/F), logic variables ($v$), and basic logic operations (negation $\neg$, conjunction $\wedge$, and disjunction $\vee$). In NLN, negation, conjunction, and disjunction are learned as three neural modules. ~\citet{leshno1993multilayer} proved that multilayer feedforward networks with non-polynomial activation can approximate any function. Thus it is possible to leverage neural modules to approximate the negation, conjunction, and disjunction operations. 

Similar to most neural models in which input variables are learned as vector representations, in our framework, T, F and all logic variables are represented as vectors of the same dimension. Formally, suppose we have a set of logic expressions $E=\{e_i\}$ and their values $Y=\{y_i\}$ (either T or F), and they are constructed by a set of variables $V=\{v_i\}$, where $|V|=n$ is the number of variables. An example logic expression is $(v_i \wedge v_j)\vee \neg v_k=T$.

\begin{figure}[htbp]
    \centering
    \vspace{-3mm}
    \includegraphics[width=1.0\linewidth]{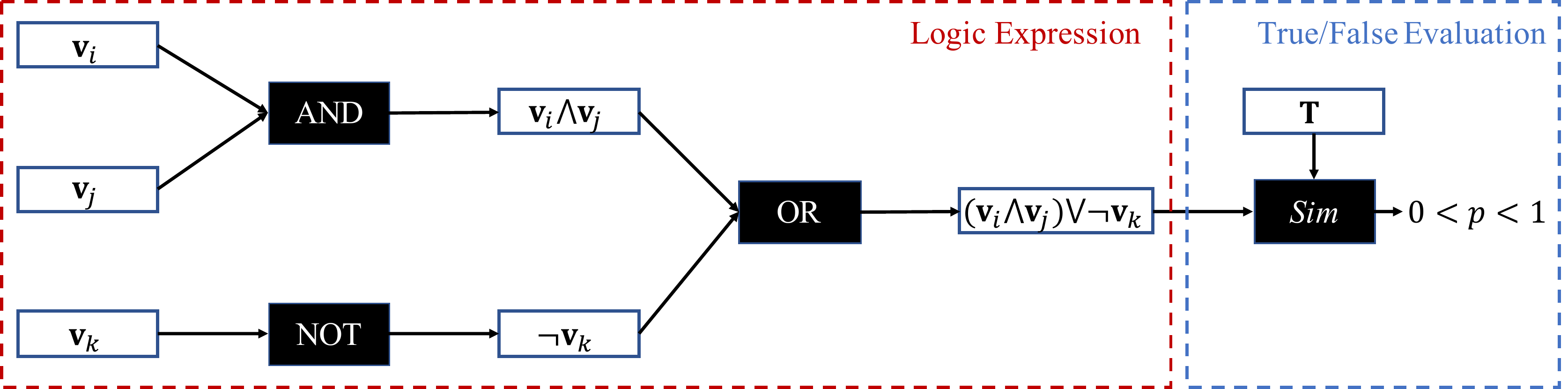}
    \vspace{-3mm}
    \caption{An example of the neural logic network.}
    \vspace{-3mm}
    \label{pic:framework}
\end{figure}

We use bold font to represent the vectors, e.g. $\mathbf{v}_i$ is the vector representation of variable $v_i$, and $\mathbf{T}$ is the vector representation of logic constant T, where the vector dimension is $d$. $\text{AND}(\cdot,\cdot)$, $\text{OR}(\cdot,\cdot)$, and $\text{NOT}(\cdot)$ are three neural modules. For example, $\text{AND}(\cdot,\cdot)$ takes two vectors $\mathbf{v}_i,\mathbf{v}_j$ as inputs, and the output $\mathbf{v}=\text{AND}(\mathbf{v}_i, \mathbf{v}_j)$ is the representation of $v_i\wedge v_j$, a vector of the same dimension $d$ as $\mathbf{v}_i$ and $\mathbf{v}_j$. The three modules can be implemented by various neural structures, as long as they have the ability to approximate the logical operations. Figure~\ref{pic:framework} is an example of the neural logic network corresponding to the expression $(v_i \wedge v_j)\vee \neg v_k$.
The red left box shows how the framework constructs a logic expression. Each intermediate vector represents part of the logic expression, and finally, we have the vector representation of the whole logic expression $\mathbf{e}=(\mathbf{v}_i \wedge \mathbf{v}_j)\vee \neg \mathbf{v}_k$. To evaluate the T/F value of the expression, we calculate the similarity between the expression vector and the $\mathbf{T}$ vector, as shown in the right blue box, where T, F are short for logic constants True and False respectively, and $\mathbf{T}$, $\mathbf{F}$ are their vector representations. Here $Sim(\cdot, \cdot)$ is also a neural module to calculate the similarity between two vectors and output a similarity value between 0 and 1. The output $p=Sim(\mathbf{e}, \mathbf{T})$ evaluates how likely NLN considers the expression to be true.

Training NLN on a set of expressions and predicting T/F values of other expressions can be considered as a classification problem, and we adopt cross-entropy loss for this task:
\begin{equation}
\label{eq:l0}
L_c = -\sum_{e_i \in E} y_i \log(p_i) + (1-y_i)\log(1-p_i)
\end{equation}

\subsection{Logical Regularization over Neural Modules}

So far, we only learned the logic operations $\text{AND}$, $\text{OR}$, $\text{NOT}$ as neural modules, but did not explicitly guarantee that these modules implement the expected logic operations. For example, any variable or expression $\mathbf{w}$ conjuncted with false should result in false $\mathbf{w}\wedge\mathbf{F}=\mathbf{F}$, and a double negation should result in itself $\neg(\neg{\mathbf{w}})=\mathbf{w}$. Here we use $\mathbf{w}$ instead of $\mathbf{v}$ in the previous section, because $\mathbf{w}$ could either be a single variable (e.g., $\mathbf{v}_i$) or an expression (e.g., $\mathbf{v}_i\wedge\mathbf{v}_j$).
A neural logic network that aims to implement logic operations should satisfy the basic logic rules. As a result, we define logic regularizers to regularize the behavior of the modules, so that they implement certain logical operations. A complete set of the logical regularizers are shown in Table~\ref{tb:laws}.
\begin{table}[htbp]
    \caption{Logical regularizers and the corresponding logical rules}
    \label{tb:laws}
    \centering
    \begin{tabular}{llll}
      \toprule
      & Logical Rule     & Equation & Logic Regularizer $r_i$\\
      \midrule
      \multirow{2}{*}{NOT} & Negation & $\neg T = F$ & $r_1=\sum_{w\in W \cup \{T\}} Sim(\text{NOT}(\mathbf{w}),\mathbf{w})$    \\ 
      & Double Negation & $\neg (\neg w) = w$ & $r_2=\sum_{w\in W} 1-Sim(\text{NOT}(\text{NOT}(\mathbf{w})),\mathbf{w})$    \\ 
      \midrule
      \multirow{4}{*}{AND} & Identity & $w \wedge T = w$ & $r_3=\sum_{w\in W} 1-Sim(\text{AND}(\mathbf{w}, \mathbf{T}),\mathbf{w})$  \\
      & Annihilator & $w \wedge F = F$  & $r_4=\sum_{w\in W} 1-Sim(\text{AND}(\mathbf{w}, \mathbf{F}),\mathbf{F})$   \\
      & Idempotence & $w \wedge w = w$  &  $r_5=\sum_{w\in W} 1-Sim(\text{AND}(\mathbf{w}, \mathbf{w}),\mathbf{w})$ \\
      & Complementation & $w \wedge \neg w = F$  & $r_6=\sum_{w\in W} 1-Sim(\text{AND}(\mathbf{w}, \text{NOT}(\mathbf{w})),\mathbf{F})$   \\
      \midrule
      \multirow{4}{*}{OR} & Identity & $w \vee F = w$ & $r_7=\sum_{w\in W} 1-Sim(\text{OR}(\mathbf{w}, \mathbf{F}),\mathbf{w})$  \\
      & Annihilator & $w \vee T = T$  & $r_8=\sum_{w\in W} 1-Sim(\text{OR}(\mathbf{w}, \mathbf{T}),\mathbf{T})$  \\
      & Idempotence & $w \vee w = w$  & $r_9=\sum_{w\in W} 1-Sim(\text{OR}(\mathbf{w}, \mathbf{w}),\mathbf{w})$ \\
      & Complementation & $w \vee \neg w = T$ & $r_{10}=\sum_{w\in W} 1-Sim(\text{OR}(\mathbf{w}, \text{NOT}(\mathbf{w})),\mathbf{T})$  \\
      \bottomrule
    \end{tabular}
\end{table}

The regularizers are categorized by the three operations. 
The equations of laws are translated into the modules and variables in our neural logic network as logical regularizers. It should be noted that these logical rules are not considered in the whole vector space $\mathbb{R}^d$, but in the vector space defined by NLN. Suppose the set of all variables as well as intermediate and final expressions observed in the training data is $W=\{w\}$, then only $\{\mathbf{w} | w\in W\}$ are taken into account when constructing the logical regularizers. Take Figure~\ref{pic:framework} as an example, the corresponding $\mathbf{w}$ in Table~\ref{tb:laws} include $\mathbf{v}_i$, $\mathbf{v}_j$, $\mathbf{v}_k$, $\mathbf{v}_i \wedge \mathbf{v}_j$, $\neg \mathbf{v}_k$ and $(\mathbf{v}_i \wedge \mathbf{v}_j) \vee \neg \mathbf{v}_k$. Logical regularizers encourage NLN to learn the neural module parameters to satisfy these laws over the variable/expression vectors involved in the model, which is much smaller than the whole vector space $\mathbb{R}^d$.

Note that in NLN the constant true vector $\mathbf{T}$ is randomly initialed and fixed during the training and testing process, which works as an indication vector in the framework that defines the true orientation. The false vector $\mathbf{F}$ is thus calculated with $\text{NOT}(\mathbf{T})$.

Finally, logical regularizers $R_{l}$ are added to the cross-entropy loss function (Eq.\eqref{eq:l0}) with weight $\lambda_{l}$:
\begin{equation}\label{eq:logical_regularizer}
L_1 = L_c + \lambda_l R_{l} = L_c
 + \lambda_{l} \sum_i r_i
\end{equation}
where $r_i$ are the logic regularizers in Table~\ref{tb:laws}.

It should be noted that except for the logical regularizers listed above, a propositional logical system should also satisfy other logical rules such as the associativity, commutativity and distributivity of AND/OR/NOT operations. To consider associativity and commutativity, the order of the variables joined by multiple conjunctions or disjunctions is randomized when training the network. For example, the network structure of $w_i \wedge w_j$ could be $\text{AND}(\mathbf{w}_i, \mathbf{w}_j)$ or $\text{AND}(\mathbf{w}_j, \mathbf{w}_i)$, and the network structure of $w_i \vee w_j \vee w_k$ could be $\text{OR}(\text{OR}(\mathbf{w}_i, \mathbf{w}_j), \mathbf{w}_k)$, $\text{OR}(\text{OR}(\mathbf{w}_i, \mathbf{w}_k), \mathbf{w}_j)$, $\text{OR}(\mathbf{w}_j, \text{OR}(\mathbf{w}_k, \mathbf{w}_i))$ and so on during training. In this way, the model is encouraged to output the same vector representation when inputs are different forms of the same expression in terms of associativity and commutativity. 

There is no explicit way to regularize the modules for other logical rules that correspond to more complex expression variants, such as distributivity and De Morgan laws. To solve the problem, we make sure that the input expressions have the same normal form -- e.g., disjunctive normal form -- because any propositional logical expression can be transformed into a Disjunctive Normal Form (DNF) or Canonical Normal Form (CNF). In this way, we can avoid the necessity to regularize the neural modules for distributivity and De Morgan laws.

\subsection{Length Regularization over Logic Variables}
We found that the vector length of logic variables as well as intermediate or final logic expressions may explode during the training process, because simply increasing the vector length results in a trivial solution for optimizing Eq.\eqref{eq:logical_regularizer}. Constraining the vector length provides more stable performance, and thus a $\ell_2$-length regularizer $R_{\ell}$ is added to the loss function with weight $\lambda_{\ell}$:
\begin{equation}
L_2 = L_c + \lambda_l R_{l} + \lambda_{\ell} R_{\ell} 
= L_c + \lambda_{l} \sum_i r_i  + \lambda_{\ell} \sum_{w\in W} \|\mathbf{w}\|_F^2
\end{equation}
Similar to the logical regularizers, $W$ here includes input variable vectors as well as all intermediate and final expression vectors.

Finally, we apply $\ell_2$-regularizer with weight $\lambda_{\Theta}$ to prevent the parameters from overfitting. Suppose $\Theta$ are all the model parameters, then the final loss function is:
\begin{equation}
  L = L_c + \lambda_l R_{l} + \lambda_{\ell} R_{\ell} + \lambda_{\Theta}R_{\Theta}
  = L_c + \lambda_{l} \sum_i r_i  + \lambda_{\ell} \sum_{w\in W} \|\mathbf{w}\|_F^2 + \lambda_{\Theta}\|\Theta\|_F^2
\end{equation}

\section{Implementation Details}

Our prototype task is defined in this way: given a number of training logical expressions and their T/F values, we train a neural logic network, and test if the model can solve the T/F value of the logic variables, and predict the value of new expressions constructed by the observed logic variables in training. We first conduct experiments on manually generated data to show that our neural logic networks have the ability to make propositional logical inference. NLN is further applied to the personalized recommendation problem to verify its performance in practical tasks.

We did not design fancy structures for different modules. Instead, some simple structures are effective enough to show the superiority of NLN. In our experiments, the $\text{AND}$ module is implemented by multi-layer perceptron (MLP) with one hidden layer:
\begin{equation}
    \text{AND}(\mathbf{w}_i, \mathbf{w}_j) = \mathbf{H}_{a2} f(\mathbf{H}_{a1} (\mathbf{w}_i|\mathbf{w}_j) + \mathbf{b}_a)
\end{equation}
where $\mathbf{H}_{a1} \in \mathbb{R}^{d\times 2d}, \mathbf{H}_{a2} \in \mathbb{R}^{d\times d}, \mathbf{b}_a \in \mathbb{R}^d$ are the parameters of the $\text{AND}$ network. $|$ means vector concatenation. $f(\cdot)$ is the activation function, and we use $relu$ in our networks. The $\text{OR}$ module is built in the same way, and the $\text{NOT}$ module is similar but with only one vector as input:
\begin{equation}
    \text{NOT}(\mathbf{w}) = \mathbf{H}_{n2} f(\mathbf{H}_{n1} \mathbf{w} + \mathbf{b}_n)
\end{equation}
where $\mathbf{H}_{n1} \in \mathbb{R}^{d\times d}, \mathbf{H}_{n2} \in \mathbb{R}^{d\times d}, \mathbf{b}_n \in \mathbb{R}^d$ are the parameters of the $\text{NOT}$ network.

The similarity module is based on the cosine similarity of two vectors. To ensure that the output is formatted between 0 and 1, we scale the cosine similarity by multiplying a value $\alpha$, following by a \textit{sigmoid} function:
\begin{equation}
    Sim(\mathbf{w}_i, \mathbf{w}_j) = sigmoid\left(\alpha \frac{\mathbf{w}_i \cdot \mathbf{w}_j}{\|\mathbf{w}_i\|\|\mathbf{w}_j\|}\right)
\end{equation}
The $\alpha$ is set to 10 in our experiments. We also tried other ways to calculate the similarity such as $sigmoid(\mathbf{w}_i \cdot \mathbf{w}_j)$ or MLP. This way provides better performance.

All the models including baselines are trained with Adam~\cite{kingma2014adam} in mini-batches at the size of 128. The learning rate is 0.001, and early-stopping is conducted according to the performance on the validation set. Models are trained at most 100 epochs. To prevent models from overfitting, we use both the $\ell_2$-regularization and dropout. The weight of $\ell_2$-regularization $\lambda_\Theta$ is set between $1\times10^{-7}$ to $1\times10^{-4}$ and dropout ratio is set to 0.2. Vector sizes of the variables in simulation data and the user/item vectors in recommendation are 64. We run the experiments with 5 different random seeds and report the average results and standard errors. Note that NLN has similar time and space complexity with baseline models and each experiment run can be finished in 6 hours (several minutes on small datasets) with a GPU (NVIDIA GeForce GTX 1080Ti).

\section{Simulated Data}
\label{sec:simulation}

We first randomly generate $n$ variables $V=\{v_i\}$, each has a value of T or F. Then these variables are used to randomly generate $m$ boolean expressions $E=\{e_i\}$ in disjunctive normal form (DNF) as the dataset. Each expression consists of 1 to 5 clauses separated by the disjunction $\vee$. Each clause consists of 1 to 5 variables or the negation of variables connected by conjunction $\wedge$. We also conducted experiments on many other fixed or variational lengths of expressions, which have similar results. The T/F values of the expressions $Y=\{y_i\}$ can be calculated according to the variables. But note that the T/F values of the variables are invisible to the model. Here are some examples of the generated expressions when $n=100$:
\begin{align*}
    (\neg v_{80} \wedge v_{56} \wedge v_{71}) \vee (\neg v_{46} \wedge \neg v_{7} \wedge v_{51} \wedge \neg v_{47} \wedge v_{26}) \vee v_{45} \vee (v_{31} \wedge v_{15} \wedge v_{2} \wedge v_{46}) = T \\
    (\neg v_{19} \wedge \neg v_{65}) \vee (v_{65} \wedge \neg v_{24} \wedge v_{9} \wedge \neg v_{83}) \vee (\neg v_{48} \wedge \neg v_{9} \wedge \neg v_{51} \wedge v_{75}) = F \\
    \neg v_{98} \vee (\neg v_{76} \wedge v_{66} \wedge v_{13}) \vee v_{97} (\wedge v_{89} \wedge v_{45} \wedge v_{83}) = T \\
    v_{43} \wedge v_{21} \wedge \neg v_{53} = F \\
\end{align*}

\subsection{Results Analysis}

\begin{table}[htbp]
\vspace{-3mm}
      \caption{Performance on simulation data}
      \label{tab:performance_sim}
      \centering
      \begin{tabular}{l|p{2.5cm}p{2.5cm}|p{2.5cm}p{2.5cm}}
        \toprule
        \multicolumn{1}{c}{}&
        \multicolumn{2}{c}{$n=1\times 10^3, m=5\times 10^3$}&
        \multicolumn{2}{c}{$n=2\times 10^4, m=5\times 10^4$}\\
        \cmidrule(lr){2-3} \cmidrule(lr){4-5}
        & Accuracy & RMSE & Accuracy & RMSE\\
        \midrule
        Bi-LSTM & 0.6128$\pm$0.0029 & 0.4952$\pm$0.0032 & 0.6826$\pm$0.0039 & 0.4529$\pm$0.0038 \\
        Bi-RNN & 0.6412$\pm$0.0014 & 0.4802$\pm$0.0033 & 0.6985$\pm$0.0023 & 0.4412$\pm$0.0005 \\
        \midrule
        NLN-$R_l$ & \textit{0.9064$\pm$0.0136} & \textit{0.2746$\pm$0.0221} & \textit{0.8400$\pm$0.0011} & \textit{0.3678$\pm$0.0013}\\
        NLN & \textbf{0.9716$\pm$0.0023}* & \textbf{0.1633$\pm$0.0080}* & \textbf{0.8827$\pm$0.0019}* & \textbf{0.3286$\pm$0.0022}*\\
      \bottomrule
    \end{tabular}
    \begin{minipage}{12cm}
      \begin{tablenotes}
      \item *. Significantly better than the other models (italic ones) with $p<0.05$
      \end{tablenotes}
    \end{minipage}
\end{table}

On simulated data, $\lambda_{l}$ and $\lambda_{\ell}$ are set to $1\times10^{-2}$ and $1\times10^{-4}$ respectively. Datasets are randomly split into the training (80\%), validation (10\%) and test (10\%) sets. The overall performances on test sets are shown on Table~\ref{tab:performance_sim}. \textbf{Bi-RNN} is bidirectional Vanilla RNN~\cite{schuster1997bidirectional} and \textbf{Bi-LSTM} is bidirectional LSTM~\cite{graves20052005}. They represent traditional neural networks. \textbf{NLN-$R_l$} is the NLN without logic regularizers. The poor performance of Bi-RNN and Bi-LSTM verifies that traditional neural networks that ignore the logical structure of expressions do not have the ability to conduct logical inference. Logical expressions are structural and have exponential combinations, which are difficult to learn by a fixed model architecture. Bi-RNN performs better than Bi-LSTM because the forget gate in LSTM may be harmful to model the variable sequence in expressions. NLN-$R_l$ provides a significant improvement over Bi-RNN and Bi-LSTM because the structure information of the logical expressions is explicitly captured by the network structure. However, the behaviors of the modules are freely trained with no logical regularization. On this simulated data and many other problems requiring logical inference, logical rules are essential to model the internal relations. With the help of logic regularizers, the modules in NLN learn to perform expected logic operations, and finally, NLN achieves the best performance and significantly outperforms NLN-$R_l$.


\begin{figure}[htbp]
  \vspace{-3mm}
  \centering
  \begin{minipage}{.45\textwidth}
    \centering
    \includegraphics[width=0.9\linewidth]{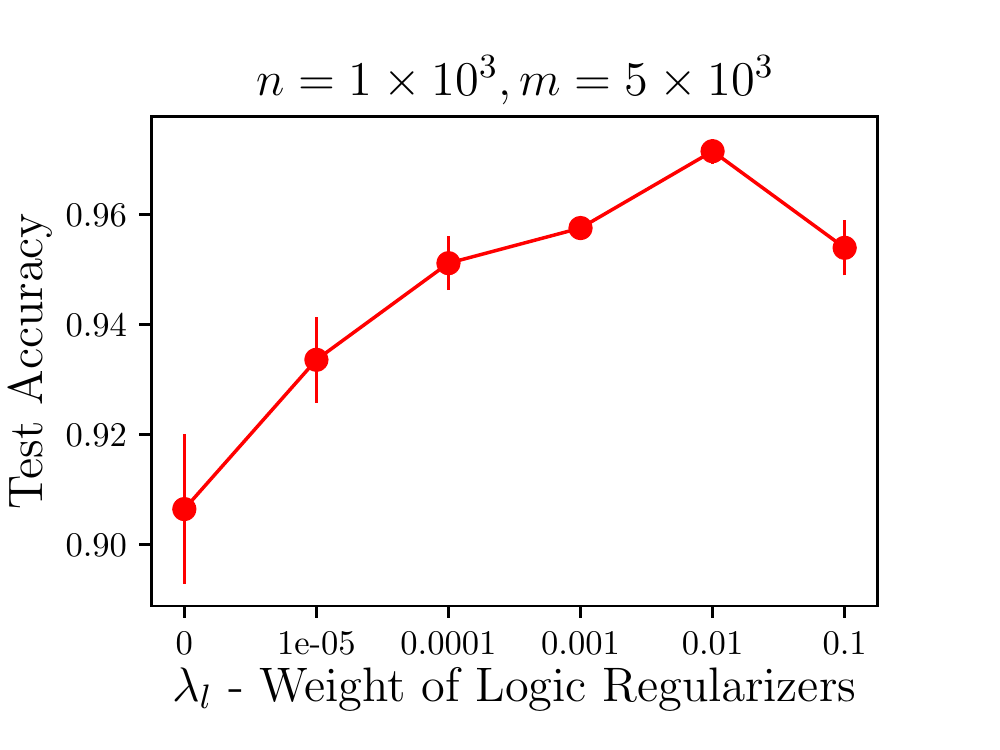}
    \vspace{-3mm}
    \caption{Performance with different weights of logical regularizers.}
    \label{fig:sim_logic_weights}
  \end{minipage}%
  \qquad
  \begin{minipage}{.45\textwidth}
    \centering
    \includegraphics[width=0.9\linewidth]{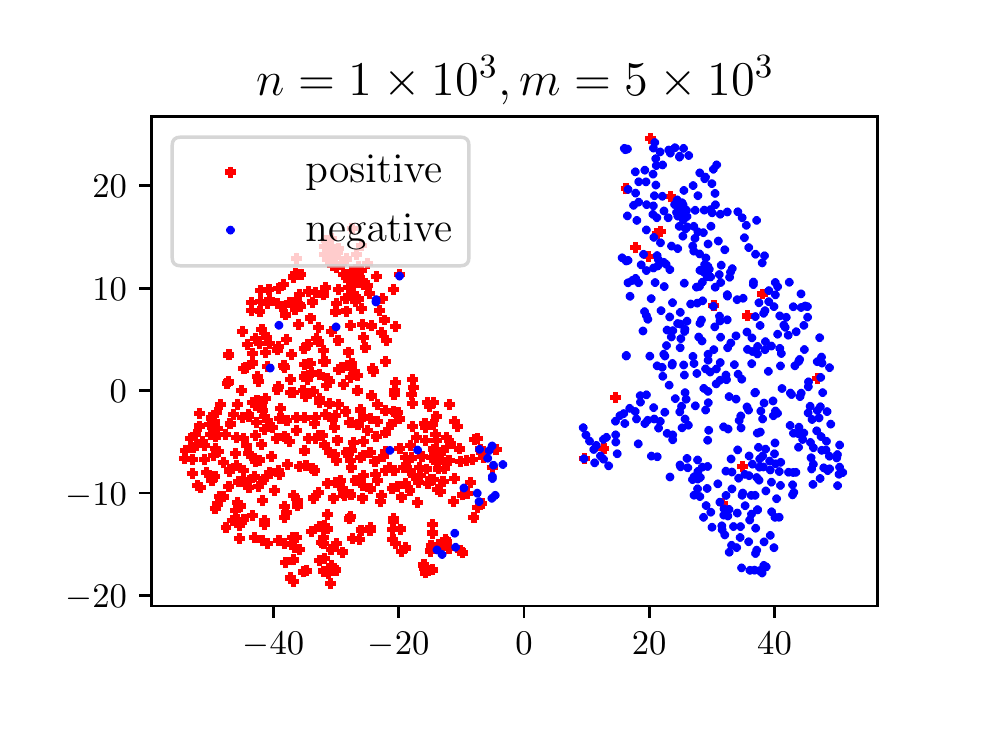}
    \vspace{-3mm}
    \caption{Visualization of the variable embeddings with t-SNE.}
    \label{fig:variable_tsne}
  \end{minipage}
\end{figure}

$\bullet$ \textit{Weight of Logical Regularizers}. To better understand the impact of logical regularizers, we test the model performance with different weights of logical regularizers, shown in Figure~\ref{fig:sim_logic_weights}. When $\lambda_l=0$ (i.e., NLN-$R_l$), the performance is not so good. As $\lambda_l$ grows, the performance gets better, which shows that logical rules of the modules are essential for logical inference. However, if $\lambda_l$ is too large it will result in a drop of performance, because the expressiveness power of the model may be significantly constrained by the logical regularizers.

$\bullet$ \textit{Visualization of Variables}. It is intuitive to study whether NLN can solve the T/F values of variables. 
To do so, we conduct t-SNE~\cite{maaten2008visualizing} to visualize the variable embeddings on a 2D plot, shown in Figure~\ref{fig:variable_tsne}. We can see that the T and F variables are clearly separated, and the accuracy of T/F values according to the two clusters is 95.9\%, which indicates high accuracy of solving variables based on NLN.

\section{Personalized Recommendation}

The key problem of recommendation is to understand the user preference according to historical interactions. Suppose we have a set of users $U=\{u_i\}$ and a set of items $V=\{v_j\}$, and the overall interaction matrix is $R=\{r_{i,j}\}_{|U|\times|V|}$. The interactions observed by the recommender system are the known values in matrix $R$. However, they are very sparse compared with the total number of $|U|\times|V|$. To recommend items to users in such a sparse setting, logical inference is important. For example, a user bought an iPhone may need an iPhone case rather than an Android data line, i.e., $\text{\textit{iPhone}} \rightarrow \text{\textit{iPhone case}} = T$, while $\text{\textit{iPhone}} \rightarrow \text{\textit{Android data line}} = F$. Let $r_{i,j}=1/0$ if user $u_i$ likes/dislikes item $v_j$. Then for a user $u_i$ with a set of interactions sorted by time $\{r_{i,j_1}=1, r_{i,j_2}=0, r_{i,j_3}=0, r_{i,j_4}=1\}$, 3 logical expressions can be generated: $v_{j_1}\rightarrow v_{j_2}=F$, $v_{j_1}\wedge \neg v_{j_2} \rightarrow v_{j_3}=F$, $v_{j_1}\wedge \neg v_{j_2} \wedge \neg v_{j_3} \rightarrow v_{j_4}=T$. Note that $a \rightarrow b = \neg a \vee b$. So in this way, we can transform all the users' interactions into logic expressions in the format of $\neg (a \wedge b \cdots) \vee c=T/F$, where inside the brackets are the interaction history and to the right of $\vee$ is the target item. Note that at most 10 previous interactions right before the target item are considered in our experiments.

Experiments are conducted on two publicly available datasets:

$\bullet$ \textbf{ML-100k}~\cite{harper2016movielens}. It is maintained by Grouplens~\footnote{\url{https://grouplens.org/datasets/movielens/100k/}}, which has been used by researchers for many years. It includes 100,000 ratings ranging from 1 to 5 from 943 users and 1,682 movies.

$\bullet$ \textbf{Amazon Electronics}~\cite{he2016ups}. Amazon Dataset~\footnote{\url{http://jmcauley.ucsd.edu/data/amazon/index.html}} is a public e-commerce dataset. It contains reviews and ratings of items given by users on Amazon, a popular e-commerce website. We use a subset in the area of Electronics, containing 1,689,188 ratings ranging from 1 to 5 from 192,403 users and 63,001 items, which is bigger and much more sparse than the ML-100k dataset. 

The ratings are transformed into 0 and 1. Ratings equal to or higher than 4 ($r_{i,j}\geq 4$) are transformed to 1, which means positive attitudes (like). Other ratings ($r_{i,j}\le 3$) are converted to 0, which means negative attitudes (dislike). Then the interactions are sorted by time and translated to logic expressions in the way mentioned above. We ensure that expressions corresponding to the earliest 5 interactions of every user are in the training sets. For those users with no more than 5 interactions, all the expressions are in the training sets. For the remaining data, the last two expressions of every user are distributed into the validation sets and test sets respectively (Test sets are preferred if there remains only one expression of the user). All the other expressions are in the training sets.

The models are evaluated on two different recommendation tasks. One is binary Preference Prediction and the other is Top-K Recommendation. The NLN on the preference prediction tasks is trained similarly as on the simulated data (Section~\ref{sec:simulation}), training on the known expressions and predicting the T/F values of the unseen expressions, with the cross-entropy loss. For top-k recommendation tasks, we use the pair-wise training strategy~\cite{rendle2009bpr} to train the model -- a commonly used training strategy in many ranking tasks -- which usually performs better than point-wise training. In detail, we use the positive interactions to train the baseline models, and use the expressions corresponding to the positive interactions to train our NLN. For each positive interaction $v^+$, we randomly sample an item the user dislikes or has never interacted with before as the negative sample $v^-$ in each epoch. Then the loss function of baseline models is:
\begin{equation}
\label{eq:bpr_loss}
   L = -\sum_{v^+} \log\big(sigmoid(p(v^+) - p(v^-))\big) + \lambda_\Theta \|\Theta\|_F^2
\end{equation}
where $p(v^+)$ and $p(v^-)$ are the predictions of $v^+$ and $v^-$, respectively, and $\lambda_\Theta \|\Theta\|_F^2$ is $\ell_2$-regularization. The loss function encourages the predictions of positive interactions to be higher than the negative samples. For our NLN, suppose the logic expression with $v^+$ as the target item is $e^+=\neg (\cdots) \vee v^+$, then the negative expression is $e^-=\neg (\cdots) \vee v^-$, which has the same history interactions to the left of $\vee$. Then the loss function of NLN is:
\begin{equation}
  L = -\sum_{e^+} \log\big(sigmoid(p(e^+) - p(e^-))\big) + \lambda_{l} \sum_i r_i  + \lambda_{\ell} \sum_{w\in W} \|\mathbf{w}\|_F^2 + \lambda_{\Theta}\|\Theta\|_F^2
\end{equation}
where $p(e^+)$ and $p(e^-)$ are the predictions of $e^+$ and $e^-$, respectively, and other parts are the logic, vector length and $\ell_2$ regularizers as mentioned in Section~\ref{sec:nln}. In top-k evaluation, we sample 100 $v^-$ for each $v^+$ and evaluate the rank of $v^+$ in these 101 candidates. This way of data partition and evaluation is usually called the Leave-One-Out setting in personalized recommendation.

\subsection{Results Analysis}

\begin{table}[htbp]
\vspace{-3mm}
  \caption{Performance on recommendation task}
  \label{tab:performance_rec}
  \centering
  \begin{tabular}{l|p{2.4cm}|p{2.4cm}|p{2.4cm}|p{2.4cm}}
    \toprule
    \multicolumn{1}{c}{}&
    \multicolumn{2}{c}{ML-100k}&
    \multicolumn{2}{c}{Amazon Electronics}\\
    \cmidrule(lr){2-3} \cmidrule(lr){4-5}
    & Preference$^1$ & Top-K$^2$ & Preference & Top-K\\
    & AUC & nDCG@10 & AUC & nDCG@10\\
    \midrule
    BiasedMF~\cite{koren2009matrix} & 0.8017$\pm$0.0002 & \textit{0.3700$\pm$0.0027} & 0.6448$\pm$0.0003 & 0.3449$\pm$0.0006 \\
    SVD++~\cite{koren2008factorization} & \textit{0.8170$\pm$0.0004} & 0.3651$\pm$0.0022 & 0.6667$\pm$0.0005 & \textit{0.3902$\pm$0.0003}\\
    NCF~\cite{he2017ncf} & 0.8063$\pm$0.0006 & 0.3589$\pm$0.0020 & \textit{0.6723$\pm$0.0008} & 0.3358$\pm$0.0011\\
    \midrule
    NLN-$R_l$ & 0.7218$\pm$0.0001 & 0.3711$\pm$0.0069 & 0.6490$\pm$0.0006 & 0.4075$\pm$0.0036\\
    NLN & \textbf{0.8211$\pm$0.0004}* & \textbf{0.3807$\pm$0.0046}* & \textbf{0.6894$\pm$0.0018}* & \textbf{0.4113$\pm$0.0015}*\\
  \bottomrule
\end{tabular}
\begin{minipage}{12cm}
  \begin{tablenotes}
  \item 1. Binary preference prediction tasks
  \item 2. Top-K recommendation tasks
  \item *. Significantly better than the best baselines (italic ones) with $p<0.05$
  \end{tablenotes}
\end{minipage}
\end{table}

On ML-100k, $\lambda_{l}$ and $\lambda_{\ell}$ are set to $1\times10^{-5}$. On Electronics, they are set to $1\times10^{-6}$ and $1\times10^{-4}$ respectively. The overall performance of models on two datasets and two tasks are on Table~\ref{tab:performance_rec}. \textbf{BiasedMF}~\cite{koren2009matrix} is a traditional recommendation method based on matrix factorization. \textbf{SVD++}~\cite{koren2008factorization} is also based on matrix factorization but it considers the history implicit interactions of users when predicting, which is one of the best traditional recommendation models. \textbf{NCF}~\cite{he2017ncf} is Neural Collaborative Filtering, which conducts collaborative filtering with a neural network, and it is one of the state-of-the-art neural recommendation models using only the user-item interaction matrix as input. Their loss functions are modified as Equation~\ref{eq:bpr_loss} in top-k recommendation tasks. 

NLN-$R_l$ provides comparable results on top-k recommendation tasks but performs relatively worse on preference prediction tasks. Binary preference prediction tasks are somehow similar to the T/F prediction task on simulated data. 
Although personalized recommendation is not a standard logical inference problem, logical inference still helps in this task, which is shown by the results -- it is clear that on both the preference prediction and the top-k recommendation tasks, NLN achieves the best performance. NLN makes more significant improvements on ML-100k because this dataset is denser that helps NLN to estimate reliable logic rules from data. Excellent performance on recommendation tasks reveals the promising potential of NLN. Note that NLN did not even use the user ID in prediction, which is usually considered important in personalized recommendation tasks. Our future work will consider making personalized recommendations with predicate logic.

\begin{figure}[htbp]
\vspace{-3mm}
  \centering
  \includegraphics[width=0.8\linewidth,trim={0 0mm 0 0mm},clip]{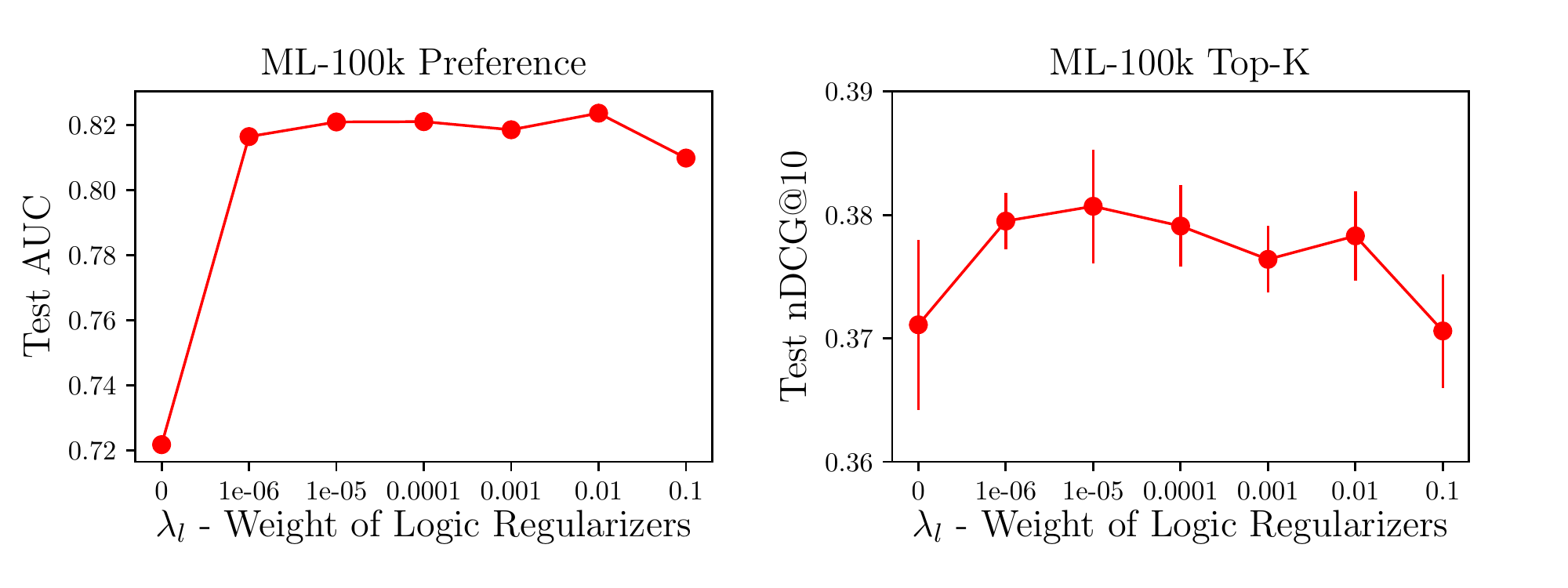}
  \vspace{-5mm}
  \caption{Performance with different weights of logic regularizers.}
  \vspace{-3mm}
  \label{fig:rec_logic_weights}
\end{figure}

$\bullet$ \textit{Weight of Logic Regularizers}. 
Results of using different weights of logical regularizers verify that logical inference is helpful in making recommendations, as shown in Figure~\ref{fig:rec_logic_weights}. Recommendation tasks can be considered as making fuzzy logical inference according to the history of users, since a user interaction with one item may imply a high probability of interacting with another item.
On the other hand, learning the representations of users and items are more complicated than solving standard logical equations, since the model should have sufficient generalization ability to cope with redundant or even conflicting input expressions. Thus NLN, an integration of logic inference and neural representation learning, performs well on the recommendation tasks. The weights of logical regularizers should be smaller than that on the simulated data because it is not a complete propositional logic inference problem, and too big logical regularization weights may limit the expressiveness power and lead to a drop in performance.

\section{Related Work}

\subsection{Neural Symbolic Learning}

\citet{mcculloch1943logical} proposed one of the first neural system for boolean logic in \citeyear{mcculloch1943logical}. Researchers further developed logical programming systems to make logical inference~\cite{garcez1999connectionist,holldobler1994towards}, and proposed neural knowledge representation and reasoning frameworks~\cite{cloete2000knowledge,browne2001connectionist} for logical reasoning. They all adopt meticulously designed neural architectures to achieve the ability of logical inference. Although ~\citet{garcez2008neural}'s framework has been verified helpful in nonclassical logic, abductive reasoning, and normative multi-agent systems, these frameworks focus more on hard logic reasoning and are short of learning representations and generalization ability compared with deep neural networks, and thus are not suitable for reasoning over large-scale, heterogeneous, and noisy data. 

\subsection{Deep Learning with Logic}

Deep learning has achieved great success in many areas. However, most of them are data-driven models without the ability of logical reasoning. Recently there are several works using deep neural networks to solve logic problems. ~\citet{hamilton2018embedding} embedded logical queries on knowledge graphs into vectors. ~\citet{johnson2017inferring} and ~\cite{yi2018neural} designed deep frameworks to generate programs and make visual reasoning automatically. ~\citet{yang2017differentiable} proposed a Neural Logic Programming system to learn probabilistic first-order logical rules for knowledge base reasoning. \citet{neural2019dong} developed Neural Logic Machines trying to learn inductive logical rules from data. Researchers are even trying to solve SAT problems with neural networks~\cite{selsam2018learning}. These works use pre-designed model structures to process different logical inputs, which is different from our NLN approach that constructs dynamic neural architectures. Although they help in logical tasks, they are less flexible in terms of model architecture, which makes them problem-specific and limits their application in a diverse range of both theoretical and practical tasks.

\section{Conclusion \& Discussions}

In this work, we proposed a Neural Logic Network (NLN) framework to make logical inference with deep neural networks. 
In particular, we learn logic variables as vector representations and logic operations as neural modules regularized by logical rules.
The integration of logical inference and neural network reveals a promising direction to design deep neural networks for both abilities of logical reasoning and generalization. 
Experiments on simulated data show that NLN works well on theoretical logical reasoning problems in terms of solving logical equations. We further apply NLN on personalized recommendation tasks effortlessly and achieved excellent performance, which reveals the prospect of NLN in terms of practical tasks. 

We believe that empowering deep neural networks with the ability of logical reasoning is essential to the next generation of deep learning. We hope that our work provides insights on developing neural networks for logical inference. In this work, we mostly focused on propositional logical reasoning with neural networks, while in the future, we will further explore predicate logic reasoning based on our neural logic network architecture, which can be easily extended by learning predicate operations as neural modules. We will also explore the possibility of encoding knowledge graph reasoning based on NLN, and applying NLN to other theoretical or practical problems such as SAT solvers.







\bibliography{neurips_2019}

\newpage
\section*{APPENDIX}

\begin{figure}[htbp]
    \centering
    \includegraphics[width=1.0\linewidth,trim={0 0mm 0 0mm},clip]{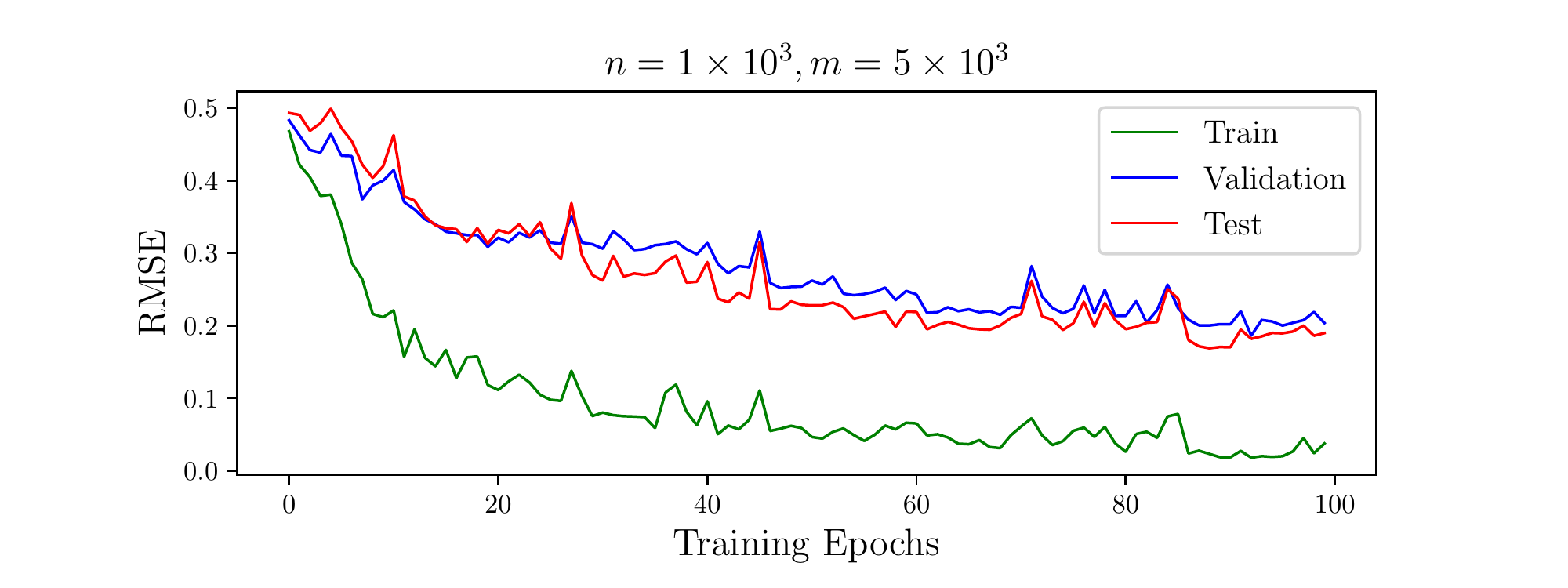}
    \vspace{-3mm}
    \caption{RMSE curves during the training process.}
    \label{fig:training_epoch_rmse}
\end{figure}

To help understand the training process, we show the curves of Training, Validation, and Testing RMSE during the training process on the simulated data in Figure~\ref{fig:training_epoch_rmse}. 

\begin{figure}[htbp]
    \centering
    \includegraphics[width=1.0\linewidth,trim={0 0mm 0 0mm},clip]{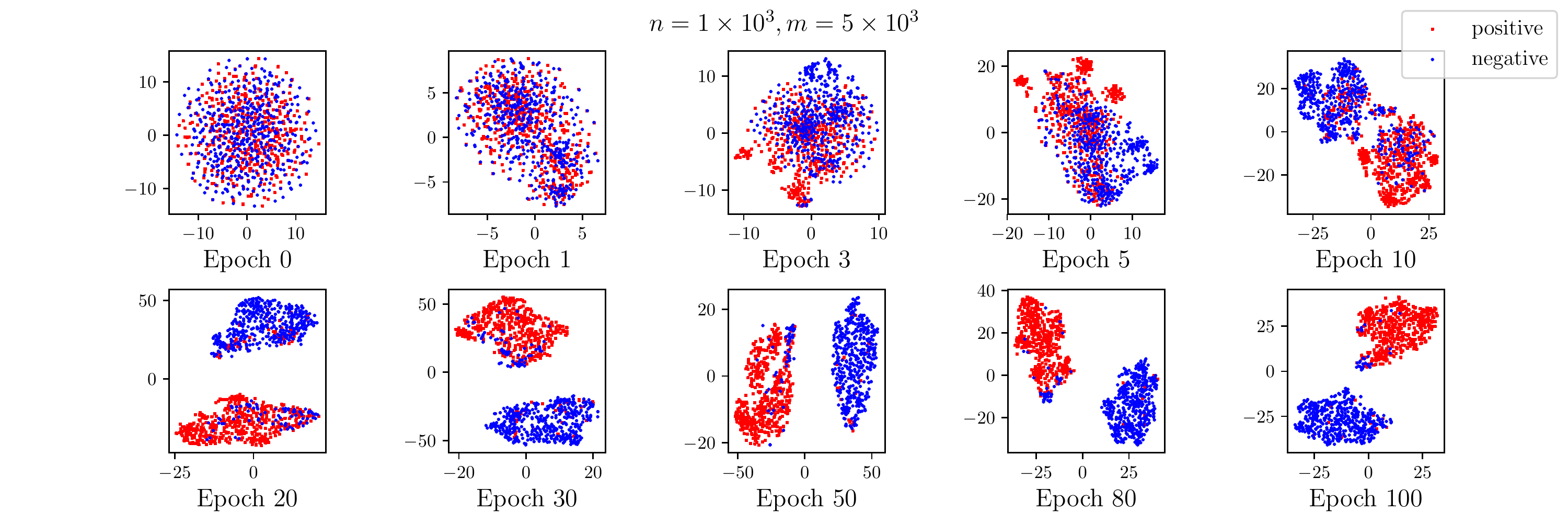}
    \vspace{-3mm}
    \caption{Visualization of the variable embeddings in different epochs based on t-SNE.}
    \label{fig:training_epoch_variable_tsne}
\end{figure}

Furthermore, the visualization of variable embeddings in different epochs are shown in Figure~\ref{fig:training_epoch_variable_tsne}.

\end{document}